\pdfoutput=1

\documentclass[11pt]{article}

\usepackage{EMNLP2022}

\usepackage{times}
\usepackage{latexsym}

\usepackage[T1]{fontenc}

\usepackage[utf8]{inputenc}

\usepackage{microtype}

\usepackage{inconsolata}

\usepackage{xspace}

\newcommand{\trusttask}{Trust Labeling Task\xspace}
\newcommand{\bert}{\textsc{Vanilla-BERT}\xspace}
\newcommand{\ctbert}{\textsc{CT-BERT}\xspace}
\newcommand{\naivebayes}{\textsc{NB}\xspace}
\newcommand{\lowtrustvax}{Low Institutional Trust\xspace}
\newcommand{\neutral}{Neutral\xspace}
\newcommand{\hightrust}{High Trust\xspace}
\newcommand{\lowtrustchat}{Low Agent Trust\xspace}
\newcommand{\viradataset}{VIRADialogs\xspace}
\newcommand{\trustdataset}{\textsc{VIRATrustData}\xspace}
\newcommand{\vira}{VIRA }

\usepackage{multirow}
\usepackage{graphicx}
\usepackage{booktabs}

\usepackage[textsize=tiny]{todonotes}

\title{VIRATrustData: A Trust-Annotated Corpus of Human-Chatbot Conversations About COVID-19 Vaccines }

\author {
Roni Friedman$^1$\thanks{\ \ These authors equally contributed to this work.}, João Sedoc$^2$\footnotemark[1], Shai Gretz$^1$\\
{\bf Assaf Toledo$^1$}, {\bf Rose Weeks$^3$}, {\bf Naor Bar-Zeev$^3$} \\{\bf Yoav Katz$^1$}, {\bf Noam Slonim$^1$}\\
 $^1$IBM Research; $^2$New York University; $^3$Johns Hopkins Bloomberg School of Public Health \\
 \
 \{roni.friedman-melamed,avishaig,katz,noams\}@il.ibm.com,
 \{assaf.toledo\}@ibm.com\\
 \{jsedoc\}@stern.nyu.edu,
 \{rweeks,nbarzee1\}@jhu.edu
}

\begin{document}
\maketitle
\begin{abstract}
Public trust in medical information is crucial for successful application of public health policies such as vaccine uptake. This is especially true when the information is offered remotely, by chatbots, which have become increasingly popular in recent years. Here, we explore the challenging task of human-bot turn-level trust classification. We rely on a recently released dataset of observationally-collected (rather than crowdsourced) dialogues with \vira chatbot, a COVID-19 Vaccine Information Resource Assistant. These dialogues are centered around questions and concerns about COVID-19 vaccines, where trust is particularly acute. We annotated $3k$ \vira system-user conversational turns for \lowtrustvax or \lowtrustchat vs. \neutral or \hightrust. We release the labeled dataset, \trustdataset, the first of its kind to the best of our knowledge. We demonstrate how modeling a user's trust is non-trivial and compare several models that predict different levels of trust.
    
\end{abstract}

\section{Introduction}
\label{sec:intro}

User's trust is a critical component of effective communication~\citep{mellinger1956interpersonal}. If dialogue systems are to become reliable and trustworthy sources of information, then modeling and understanding interlocutor trust is paramount. While there are a plethora of conversational agents for public health information about COVID-19, there have been no observational studies of user trust.  

Turn-level user trust evaluation can be utilized in various ways. First, during the dialogue, it can be used to adjust a dialogue system's responses or elicit human intervention. Post dialogue, it can (i) Assist in identifying  topics or particular dialogue system responses inducing mistrust or improving trust; and (ii) Provide insight into the profile of people chatting with that certain dialogue system, that can drive improvement in the dialogue system's design.

Importantly, trust is \textit{not} sentiment. This may seem obvious, but negative sentiment towards COVID-19 is often opposed to 
vaccine hesitancy (hereafter, \lowtrustvax). For instance, 
\textit{I'm young and healthy, I don't need the vaccine} presents a positive sentiment overall and a negative sentiment towards the vaccine, mapping to \lowtrustvax. \textit{Where can I get a vaccine?} shows no explicit sentiment or stance toward vaccines, yet distinctly conveys high trust in the vaccine.

In this work, we explore the challenging task of human-agent turn-level trust classification, in the domain of COVID-19 vaccine-related conversations. We rely on \textit{real} public dialogue system data, which is rarely available to the community due to the difficulty in collecting it and to privacy limitations in making it public. 

The data, \viradataset \cite{system_paper}, contains $8$k conversations of actual users with \vira -- COVID-19 Vaccine Information Resource Assistant -- a dialogue system which consults in a domain where trust is particularly acute. We annotated a subset of $3$k \vira system-user conversational turns for \lowtrustvax (in context of the \emph{vaccine}) or \lowtrustchat (towards the \emph{dialogue system}) vs \neutral or \hightrust.

We refer to this dataset as \trustdataset and make it public as part of this paper.\footnote{\url{https://research.ibm.com/haifa/dept/vst/debating_data.shtml}} Then we demonstrate the utility of these annotations for building predictive models of trust.

\section{Background \& Related Work}
\label{sec:related-work}

\subsection{What is trust?}

Trust is multidimensional and different aspects of trust should be distinguished. 
We can decompose interpersonal trust into competence and benevolence, then these can be broken down further into more factors: competence, expertness, dynamism, predictability, goodwill/morality, goodwillness/intentions, benevolent/caring/concern, responsiveness, honesty, credibility, reliability, openness/openmindedness, careful/safe, shared understanding, personal attractiveness~\citep{mcknight1996meanings}. Furthermore, we may also restate this as trust in intentions vs trust in beliefs.  

The aforementioned definitions are mostly interpersonal (or in our case human-bot) relations; however, institutional trust is another construct which relates to a larger third party~\citep{mcknight1996meanings,harrison2001trust,watson2005can}. Here, institutional trust is the assumed (lack of) benevolence of institutions and (lack of) reliability of COVID vaccines. 

\subsection{Trust \& Dialogue Systems}

In the human-computer interaction (HCI) community there is work on human-computer trust (HCT)~\citep{madsen2000measuring,sebo2019don,gebhard2021modeling}. Notably, \citet{madsen2000measuring} make a 
similar distinction to ours, 
of trust towards the agent (micro trust) and institutional trust (macro-trust). 
This work then relates to a simpler measure of trust via desirability or perceived shared emotions (a.k.a. empathy)~\citep{kraus2021towards}. 

Another active area of research is the language of chat systems that enhance user system trust~\citep{gebhard2021modeling,zhou2019trusting,lukin-etal-2018-consequences}, 
for instance the analysis of  humor~\citep{ritschel-andre-2018-shaping}. 
Yet others have analyzed trust for customer service chatbots~\citep{schanke2021estimating}. Our work instead focuses on annotating and modeling micro and macro trust on real observational data.

\section{\trustdataset Dataset}

To create \trustdataset, we annotated a subset of $3$k user responses in \viradataset for institutional and agent trust level. Next, 
we describe the process of creating this dataset. We release \trustdataset to the research community.

\subsection{Data Selection and Pre-processing}

First, we 
determined 
which user responses to label from \viradataset, as part of the \trusttask. We randomly sampled user responses under the following limitations:

    (1) Each dialogue in \viradataset contributed at most 
    one turn. 
    (2) Dialogues containing at least one user response marked with the \emph{is\_profanity} - indicating a toxic comment - were excluded. 
    (3) Only user responses between 2 and 200 characters were included. 
    (4) Only user responses containing alphabet letters were included.

As described in \citet{system_paper}, user responses in \viradataset were modified to mask personal user information as well as toxic words. 
In addition, to facilitate annotators work, 
all occurrences of 
\vira were replaced by 'chatbot'.

\subsection{Trust Annotations Collection}
To label
user responses for trust, we conducted a crowd annotation task using the Appen platform.\footnote{\url{http://appen.com/}} Annotators were presented with a single dialogue turn each time, consisting of a system message 
(for context), 
followed by a user response.

Annotators were asked 
two questions, directed at the user response. 
\textit{Question I} was aimed at determining the perceived level of trust: 
\begin{quote} 
(I) What is the trust level reflected by the user response? [Options: Low trust, High trust, Not sure/Hard to tell]
\end{quote}
\textit{Question II} was a conditionally forking follow up to Question I (see Figures \ref{fig:task1}, \ref{fig:task2} in Appendix). If a worker answered Low or High trust in Question I, Question II assessed the target of this perceived trust/mistrust:
\begin{quote} 
(II.a) What is the main target of the user's [trust/mistrust]? 
[Options: (i) The vaccine or related institutions/people; (ii) The chatbot]
\end{quote}
To avoid annotator bias towards an answer that is not followed by an
additional question, we included a follow up question if a worker 
first answered 
Not sure/Hard to tell:
\begin{quote} 
(II.b) Does the user express any kind of sentiment? [Options: Yes, No]
\end{quote} 

Annotators were provided with examples for different possible answers (see Figure \ref{fig:guidelines} in Appendix), as well as hidden embedded test questions, based on presumed ground truth. These question alerted annotators when they failed on them, 
hence provided 
additional feedback on task expectation 
while monitoring 
annotator quality, as detailed in Section~\ref{sec:qualitycontrol}. 

Annotators were alerted that the task data may include toxic comments 
as well as misleading assumptions regarding COVID-19 vaccines. 
Each turn was annotated by $7$ annotators. Annotators received $\$9$/h on average, which is higher than the US minimum wage ($\$7.25$).

\subsection{Quality control}
\label{sec:qualitycontrol}
We employed the following measures for quality control:

    (1) \textit{Test Questions} - $25\%$ of the questions answered by the annotators were hidden test questions based on ground truth. Annotators failing more than $30\%$ of them were removed from the task and their annotations were discarded.
    
    (2) \textit{Kappa Analysis} - Following \citet{toledo-etal-2019-automatic}, we calculated (I) \textit{Annotator-}$\kappa$: Pairwise Cohen's kappa ($\kappa$) \cite{cohen1960} for each pair of annotators sharing at least $50$ common judgements.
    We then averaged all pairwise $\kappa$ for each annotator having at least $5$ such pairwise $\kappa$ values estimated. Annotations of annotators with \textit{Annotator-$\kappa$} below $0.35$ were discarded. 
    (II) \textit{Task-Average-$\kappa$}: Obtained by averaging \textit{Annotator-$\kappa$} and is used to monitor task quality. 
    
    (3) \textit{Selected crowd annotators} - Following \citet{Gretz2020}, the task was available to a selected group of 
    around 
    $600$ annotators who performed well on past tasks of our team.
    
Overall \trusttask \textit{Task-Average-$\kappa$} on was $0.54$ and $0.48$ on Question I and Question II, respectively,
which is reasonable
for such subjective tasks (e.g., \citet{corpus_wide_arg_mining}).

\begin{table}
\resizebox{\columnwidth}{!}{%
\begin{tabular}{|l|l|l|}
\hline
\textbf{Label}         & \textbf{Ratio} & 
\textbf{User Response Example}                                \\ \hline
\neutral  & $62\%$   & Should I wear a mask indoors?   \\ \hline
\lowtrustvax & $25\%$   & the vaccine is a tool for government control\\ \hline
\hightrust     & $12\%$   & What is the best vaccine for me?              \\ \hline
\lowtrustchat  & $1\%$    &    Why are you spreading misinformation?\\ \hline
\end{tabular}%
}
\caption{\trustdataset class distribution and  examples.}
\label{tab:label-ratio-examples}
\end{table}

\subsection{Post processing}
To construct \trustdataset from the annotations we collected, we filtered labels as follows:

(1) We only included turns in which at least $60\%$ of annotators agreed on the majority label for Question I; 
(2) We further discarded turns in which the majority label for \textit{Question I} was \textit{Low Trust}, but no annotator majority was established for \textit{Question II} label. 

We define $4$ classes for \trustdataset: 

    \textbf{\neutral} -  majority judgement on Question I was \textit{Not sure/Hard to tell}; 
    
    \textbf{\hightrust} -  majority judgement on Question I was \textit{High Trust}\footnote{We do not further split the category upon Question II, as the answer was almost uniformly \textit{The vaccine or related institutions/people}. }; 
    
    \textbf{\lowtrustvax} -  majority judgement on Question I was \textit{Low Trust} and for Question II was \textit{The vaccine or related institutions/people}; 
    
    \textbf{\lowtrustchat} -  majority judgement on Question I was \textit{Low Trust} and for Question II was \textit{The Chatbot}.

Our data collection yielded $3{,}025$ fully labeled system-user turns. The distribution of the classes as well as examples for them can be found in Table~\ref{tab:label-ratio-examples}.

\begin{table}
\small
\begin{center}
\begin{tabular}{ |p{3.75cm}|p{3.2cm}| }
\hline
 \textbf{Class} & \textbf{Lemmas} \\
 \hline
 \neutral & \{\textit{delta, variant, immunity}\}\\
 \hline
\lowtrustvax & \{\textit{die, side, cause}\}\\
 \hline
 \hightrust & \{\textit{booster, get, shoot}\}\\
 \hline
\lowtrustchat & \{\textit{you, vaccine, need}\}\\
 \hline
 \end{tabular}
 \end{center}
 \caption{Top lemmas for each class in \trustdataset.}
\label{tab:info_gain_new}
\end{table}

\subsection{Data analysis}

To examine the lexical characteristics of each of the $4$ classes, we performed information-gain analysis. 
First, we lemmatized user responses, and kept lemmas with a word-frequency ($wf$) of at least $20$ in each class. We then calculated the Kullback–Leibler ($kl$) divergence between lemma distribution over classes and 
prior classes distribution in the train set, and ranked the lemmas by their $wf*kl$ score.

Table~\ref{tab:info_gain_new} presents the top ranked lemmas for each class, demonstrating the differences in class content. User responses labeled as \neutral for trust revolve around COVID-19 rather than the vaccines, with questions about variants and general immunity. \lowtrustvax labeled responses discuss death (from the vaccine), side effects, and other harms that can be \textit{caused} by the vaccine; whereas \hightrust responses focus on boosters/shots (here in the verb form `shoot') and where and when to \textit{get} them.

\section{Experiments}

\begin{table*}[t]
\resizebox{\textwidth}{!}{%
\begin{tabular}{|l|lll|ll|ll|ll|ll|}
\hline
\multirow{2}{*}{Model} &
  \multicolumn{3}{c|}{overall} &
  \multicolumn{2}{c|}{Low Inst. Trust} &
  \multicolumn{2}{c|}{\lowtrustchat} &
  \multicolumn{2}{c|}{\neutral} &
  \multicolumn{2}{c|}{\hightrust} \\ \cline{2-12} 
 &
  \multicolumn{1}{l|}{acc} &
  \multicolumn{1}{l|}{mac-F1} &
  w-F1 &
  \multicolumn{1}{l|}{prec} &
  recall &
  \multicolumn{1}{l|}{prec} &
  recall &
  \multicolumn{1}{l|}{prec} &
  recall &
  \multicolumn{1}{l|}{prec} &
  recall \\ \hline
\bert &
  \multicolumn{1}{l|}{0.873} &
  \multicolumn{1}{l|}{0.716} &
  0.873 &
  \multicolumn{1}{l|}{0.840} &
  0.823 &
  \multicolumn{1}{l|}{0.350} &
  0.222 &
  \multicolumn{1}{l|}{0.893} &
  0.911 &
  \multicolumn{1}{l|}{0.868} &
  \textbf{0.848} \\ \hline
\ctbert &
  \multicolumn{1}{l|}{\textbf{0.898}} &
  \multicolumn{1}{l|}{\textbf{0.762}} &
  \textbf{0.888} &
  \multicolumn{1}{l|}{\textbf{0.846}} &
  \textbf{0.883} &
  \multicolumn{1}{l|}{0.650} &
  \textbf{0.311} &
  \multicolumn{1}{l|}{\textbf{0.910}} &
  \textbf{0.915} &
  \multicolumn{1}{l|}{\textbf{0.892}} &
  0.816 \\ \hline
\naivebayes &
  \multicolumn{1}{l|}{0.766} &
  \multicolumn{1}{l|}{0.573} &
  0.754 &
  \multicolumn{1}{l|}{0.751} &
  0.596 &
  \multicolumn{1}{l|}{\textbf{1.000}} &
  0.111 &
  \multicolumn{1}{l|}{0.769} &
  0.901 &
  \multicolumn{1}{l|}{0.783} &
  0.482 \\ \hline
\end{tabular}%
}
\caption{Evaluation of baselines over \trustdataset. mac-F1 stands for macro F1 avg. and w-F1 stands for weighted macro F1 avg.}
\label{tab:baseline-scores-2}
\end{table*}

To report baseline results, we split the $3025$ turns in \trustdataset to $1816$ for training, $301$ for dev (used by neural models for early stopping) and $908$ for testing, while preserving class distribution in all sets.
For predicting trust-level we only used the user utterances. 

\subsection{Baselines Algorithms}

We evaluated the following baselines:

\textbf{\naivebayes} - Multinomial Naive Bayes, based on word count vectors.\footnote{\url{https://scikit-learn.org/stable/modules/generated/sklearn.naive_bayes.MultinomialNB.html}}

\textbf{\bert} - BERT-Large-uncased \cite{vanilla_bert}, fine-tuned on the training set.

\textbf{\ctbert} \cite{ctbert} - A BERT-Large  model pre-trained on 97M messages from twitter about COVID-19, fine-tuned on the training set.

The implementation details of both \bert and \ctbert are in Appendix~\ref{sec:model_implementation}.

\subsection{Results}
Results are presented in Table \ref{tab:baseline-scores-2}.\footnote{For \bert and \ctbert we show average metrics over $5$ random seeds.} \ctbert performed best overall, 
as well as in the two most common classes, \lowtrustvax and \neutral classes,
demonstrating the advantage of 
domain adaptation. 
On \hightrust, \bert provided the best recall, albeit with lower precision. All models struggled to detect \lowtrustchat, presumably as it is infrequent in the training set.

\subsection{Error analysis}
We reviewed $48$ cases where all baselines predicted the labels incorrectly.\footnote{In \bert and \ctbert - wrong in all or 
in majority of the $5$ runs of each.} 

We identified 6 ``incorrectly'' labeled cases, due to borderline interpretation, such as ``how safe are the vaccines'' labeled as \lowtrustvax. 

Often, however, models failed on terms that 
required subtle context to disambiguate. E.g.,  
\textit{Immune system} 
was common in the \lowtrustvax train set,  
but also in other classes, in the context of a compromised immune system and a question regarding a person that trusts their immune system. 
As such, ``My immune system can deal with covid 19'' is labeled by annotators as \lowtrustvax, but is generally predicted as \neutral by the model.  
Other examples included are \neutral instances such as ``Are there side effects'' or ``side effects Pfizer'', predicted as \lowtrustvax by all models. However, these differ mostly in tone from true \lowtrustvax examples (e.g., ``Isn't the vaccine unsafe because of side effects?''). 

\lowtrustchat present is 
a different issue - it was under-predicted by all models due its low prevalence in the data. This was despite the fact that it is a relatively well defined class, usually containing direct reference to the dialogue system (e.g., ``You are not answering my question...'', ``who pays you?'' etc.). 

\section{Conclusions}
In this paper we highlight the need to detect Institutional/Agent low trust, reflected by users of public health chatbots.
We implemented a corresponding crowd sourced annotation task, on top of \viradataset  --  
a recently released dataset of real-world human-bot dialogues around COVID-19 vaccines. We share the \trustdataset dataset, along with baseline results of several algorithms. We hope that this resource will be valuable to relevant research communities. 

Future work should collect similar labeled data on domains beyond COVID-19 vaccines to support the development of more advanced models that detect Institutional/Agent Low Trust reflected by users.
\section*{Limitations}

The dataset released in this paper presents a few limitations. 
\begin{itemize}
    \item {The dataset covers a single domain, Covid-19 vaccine mistrust, with related unique attributes. Applying trust detection to other domains requires further data collection.}
    \item {The dataset was collected over months, hence it may have specific linguistic characteristics associated with this time period.}
    \item{Agent mistrust is rare in the given settings and therefore, s under represented.}
    \item{Given that the annotation context for each input was a single dialogue turn, trust level may not always have been clear, and this might have led to an increase in the \neutral class ground truth. Notably, \hightrust class ground truth is limited to inputs that clearly indicate a vaccinated user or a wish to be vaccinated}.
\end{itemize}

\section*{Acknowledgements}

We would like to thank Pooja Sangha and Jae Hyoung Lee for their help in defining the trust categories.
\bibliography{anthology,custom}
\bibliographystyle{acl_natbib}
\appendix

\section{Model Implementation Details}
\label{sec:model_implementation}

For both \bert and \ctbert we use AdamW optimizer with a learning rate of $3$e-$5$ and a batch size of $32$. We fine-tune the model for $6$ epochs and select the best performing checkpoint on the dev set according to overall accuracy. For \ctbert we used COVID-Twitter-BERT v2.\footnote{\url{https://huggingface.co/digitalepidemiologylab/covid-twitter-bert-v2}}

\noindent
\begin{figure}[htb]
\begin{center}
\includegraphics[width=\columnwidth]{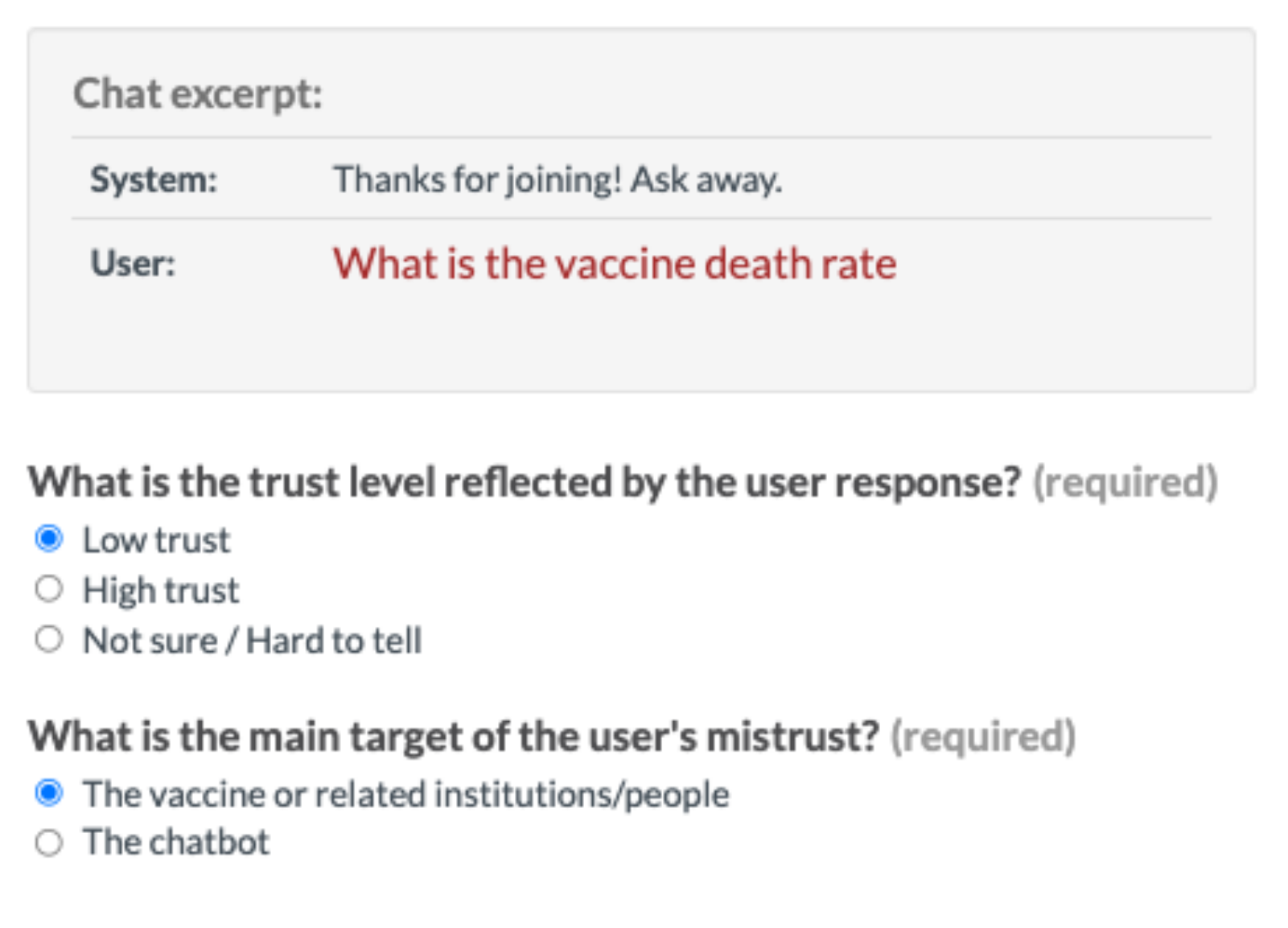}
\caption{\trusttask interface, \textit{Low trust} selection}
\label{fig:task1}
\end{center}
\end{figure}

\noindent
\begin{figure}[htb]
\begin{center}
\includegraphics[width=\columnwidth]{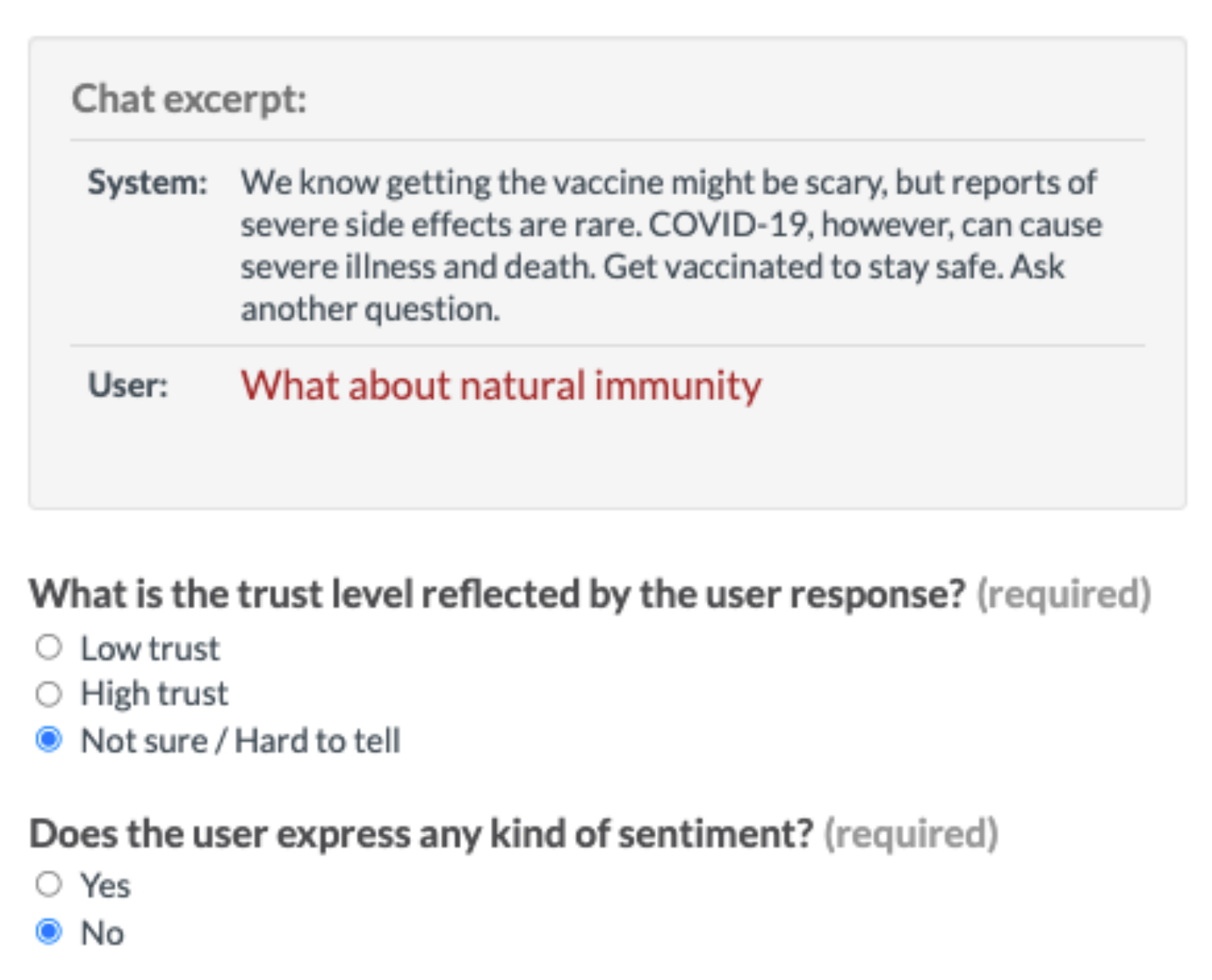}
\caption{\trusttask interface, \textit{Not suer/Hard to tell} selection}
\label{fig:task2}
\end{center}
\end{figure}

\noindent
\begin{figure}[htb]
\begin{center}
\includegraphics[width=\columnwidth]{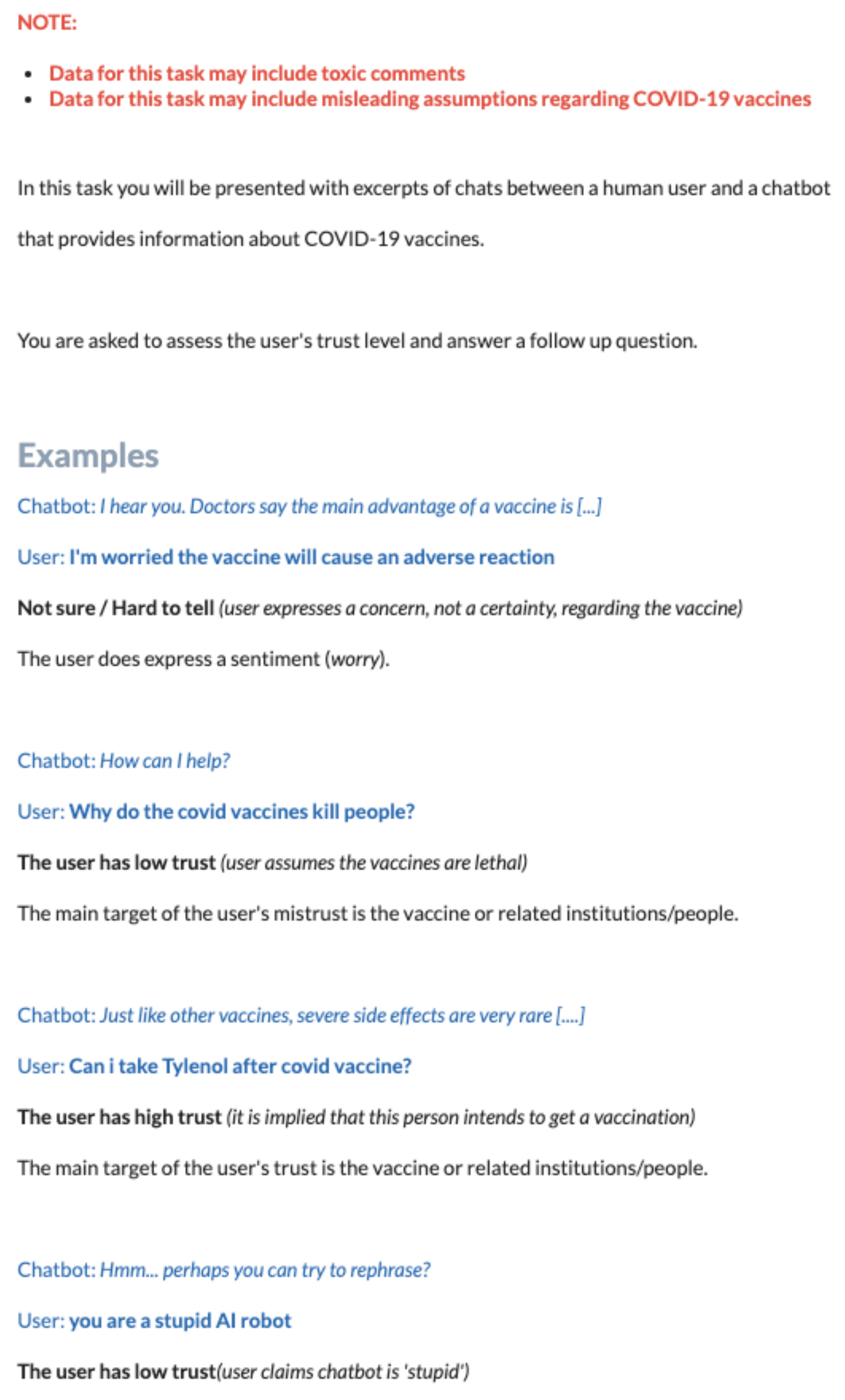}
\caption{\trusttask Guidelines}
\label{fig:guidelines}
\end{center}
\end{figure}

\end{document}